# Conditioning on disjunctive knowledge: defaults and probabilities

Eric Neufeld* and J.D. Horton†
School of Computer Science,
University of New Brunswick,
Fredericton, New Brunswick, CANADA, E3B 5A3

## Abstract

Many writers have observed that default logics appear to contain the "lottery paradox" of probability theory. This arises when a default "proof by contradiction" lets us conclude that a typical $X$ is not a $Y$ where $Y$ is an unusual subclass of $X$.

We show that there is a similar problem with default "proof by cases" and construct a setting where we might draw a different conclusion knowing a disjunction than we would knowing any particular disjunct. Though Reiter's original formalism is capable of representing this distinction, other approaches are not. To represent and reason about this case, default logicians must specify how a "typical" individual is selected.

The problem is closely related to Simpson's paradox of probability theory. If we accept a simple probabilistic account of defaults based on the notion that one proposition may *favour* or increase belief in another, the "multiple extension problem" for both conjunctive and disjunctive knowledge vanishes.

*Research supported by Natural Science and Engineering Research Council of Canada grants OGP0041937 and EQP0041938.
†Research supported by Natural Science and Engineering Research Council of Canada grant OGP0005376.

## 1 Introduction

The idea that intelligence, artificial or otherwise, involves the ability to "jump" to "default" conclusions is an attractive one; if true, it would explain a lot of intelligent activity without the need for numeric probability distributions. The classic example is the "birds fly" problem. Given that some individual *tweety* is a bird, we "jump" to the conclusion she flies. When we later discover she is an *emu*, we *retract* that *defeasible* conclusion and decide instead that she doesn't fly.

Reiter's [1980] formalism can represent this in two different ways. One way is as follows:

$$\frac{bird : M fly}{fly}$$

$$\frac{emu : M \neg fly}{\neg fly}$$

$$emu \supset bird.$$

The first default is read as follows: if *bird* is true, and it is consistent to assume *fly*, then infer *fly*. We say this representation is in *prerequisite form*, since every default has a prerequisite, following the terminology of Etherington [1987]. With the aid of "theory comparators" such as "specificity" [Poole,1985] or "inferential distance" [Touretzky,1984], it is possible to conclude that if Polly is an emu, then Polly doesn't fly. If we only know Polly is a bird, then we conclude that Polly can fly, but we can conclude nothing else about Polly.

We can also represent the knowledge as follows:
$$\frac{: M\,bird \to fly}{bird \to fly}$$
$$\frac{: M\,emu \to \neg fly}{emu \to \neg fly}$$
$$emu \supset fly.$$

This is in *consequent* form. This representation is closely related to the system of Poole and his colleagues [Poole *et al*,1987,Poole,1988] and it lets us consistently show that if we have observed a *bird*, that bird is not an *emu* using the contrapositive form of the second default.

This doesn't seem unreasonable; we can give a default "proof by contradiction" that birds are (typically) not emus: if birds were emus, then birds wouldn't (typically) fly. But birds *do* (typically) fly; a contradiction.[1]

Although no one argues that *emus* aren't rare, Poole [1987] notes that this leads to a questionable side effect sometimes called the "dingo paradox". In [Neufeld,1988] this is discussed in the context of the lottery paradox of probability theory [Kyburg,1988] and it is described as a variant of that paradox in [Neufeld and Poole,1988,Neufeld,1989a,1989b]. Kyburg [1988] states that the nonmonotonic logic formalisms contain the lottery paradox; and this is stated from within the camp of default logic in [Poole,1989].

In this paper, we question the idea of default "proof by cases". Intuition suggests that if $a$ typically implies $c$ and $b$ typically implies $c$, then $a \vee b$ typically implies $c$. Suppose, however, we give to "$a$ typically implies $b$" the probabilistic interpretation $p(b|a) > p(b)$. This is the weakest probabilistic property we believe a default ought to have, *whatever else a default may mean*. We will say *a favours b* when this is true, following [Chung,1943].[2] It then becomes simple to construct a counterexample to the notion of default "proof by cases" and multiple extensions arise for both conjunctive and disjunctive knowledge. We conclude with discussion of the implications.

## 2 Does an emu or ostrich run?

Poole [1988] poses the following. Suppose emus (typically) run and ostriches (typically) run. We can write this in prerequisite form as
$$\frac{ostrich : M\,run}{run}$$
$$\frac{emu : M\,run}{run}.$$

Poole [1988] observes that if we know only that Polly is an emu or an ostrich, but do not know which, we cannot conclude that Polly runs. We simulate his system by rewriting the defaults in consequent form:
$$\frac{: M\,ostrich \to run}{ostrich \to run}$$
$$\frac{: M\,emu \to run}{emu \to run},$$
which allows a default "proof by cases" of *run* given *emu* ∨ *ostrich*. The idea that this should be the case appears also in Delgrande's [1987] logic NP which contains the axiom schema

$$(a \Rightarrow c \wedge b \Rightarrow c) \supset (a \vee b \Rightarrow c)$$

where $\Rightarrow$ is Delgrande's "variable conditional" operator. Similarly, Geffner's [1988] system has the inference rule

---

[1] Formally, it is possible to consistently assume both consequent form defaults and thus derive ¬*emu* from *bird*.

[2] In another work [Neufeld and Poole,1988], we call this "confirmation". Wellman [1987] pursues the same idea as "qualitative influence" in the realms of planning and diagnosis.

$$\text{If } \Gamma, H' \mathrel{|\!\sim} H \text{ and } \Gamma, H'' \mathrel{|\!\sim} H, \text{ then}$$
$$\Gamma, H' \vee H'' \mathrel{|\!\sim} H,$$

where $\mathrel{|\!\sim}$ is Geffner's provability operator.

Poole [1988] argues that prerequisite form gives "unintuitive" results. We agree with his particular example, but argue that the different representations point to a variation of the "multiple extension problem" [Hanks and McDermott,1986]. It is a premise of nonmonotonic logic that it should be possible to make a different inference from $a \wedge b$ than from either conjunct. We ask the same question about *disjunctive* knowledge: do we ever want to draw a different conclusion from $a \vee b$ than from either disjunct?

Certainly we can *write down* such a set of defaults; can we provide a semantic account for doing so? In the next section, we describe a probabilistically motivated counterexample to this intuition. If our motivation is correct, we believe at least one of the following must be true:

1. Reiter's *formalism* does not give "unintuitive" results, but rather, a default reasoner must know when it is making inferences on the basis of knowing only a disjunction rather than knowing one of the disjuncts.

2. Those developing default logics must provide a more rigorous account of what is meant by "typically".

3. The proper formalism for reasoning under uncertainty, even when numeric probability distributions are unavailable, is standard probability theory.

## 3 Arts students and science students

Let $c_i$ mean that an individual is a student in a class $i$; let $a$ mean the student is an arts student and let $s$ mean the student is a science student. Ignoring the issue of prerequisite form, consider a set of defaults of the form

$$\frac{c_i : Ms}{s}$$

that intuitively means a student in class $i$ is (typically) a science ($s$) student. Consider the following scenario. Suppose there are only two classes under consideration and both have three science students and two arts students. Because they are core courses, they have the same science students but different arts students. Finally, assume there is one more science student in the domain.

Suppose we interpret this default to mean "favours", i.e., $p(s|c_i) > p(s)$. The reader can easily verify that if such inequalities are a partial account of defaults, there is also a need to condition differently on the disjunction than on the disjuncts. (This turns out to be true if we interpret defaults to mean "most" [Bacchus,1989]).

There are some straightforward arguments against this.

### 3.1 "We choose a student from the class first"

A reasonable argument is this: if we enter the *any* class $c_i$, the typical student will be a science student.

This argument does not allow us to represent the different ways we might select a typical student. If we enter $c_1$, but don't know which class we are in, that class favours the conclusion that the typical student is a science student. But this is not the only way we might meet someone in one of those classes: if the students of the $c_i$ have banded together to complain that the courses were too technical (for example), we might suspect that the typical member of such a group is an arts student, even though we know only that the student is a member of the disjunction of the classes.

This is the heart of the problem: how is the "typical" student in $c_1 \vee c_2$ selected? Do we want to know whether $favours(s,c_1) \vee favours(s,c_2)$ or $favours(s, c_1 \vee c_2)$ is true?

Note that Poole's "running emus" is the special case where conditioning on either disjunct yields the same probabilistic answers as conditioning on the disjunction. This is straightforward to prove since *emu* and *ostrich* are mutually exclusive [Neufeld,1989b].

**Proposition 1** Let $a$ and $b$ be mutually exclusive and separately favoured by $c$. Then $a \vee b$ is favoured by $c$.
**Proof:** From the premises, $p(ab) = p(ab|c) = 0$. From the disjunction rule

$$p(a \vee b|c) = p(a|c) + p(b|c)$$

and

$$p(a \vee b) = p(a) + p(b).$$

Both quantities on the right hand side of the first equality are greater than the respective quantities on the right hand side of the second and the desired inequality follows. □

### 3.2 "The probability is close to 1/2 and is unininteresting"

We have been told that the probabilities involved are too close to 1/2 to be interesting. It is easy for the *conjunctive* case to construct sets $a$, $b$ and $c$ so that for arbitrary probability values $v_1, v_2$ in the open unit interval, $p(c|a) = p(c|b) = v_1$ and $p(c|ab) = v_2$.

To achieve a similar result for the *disjunctive* case, we need only create enough disjuncts. Returning to the "arts and science" example, suppose we want $p(s|c_i)$ to be at least $v_1$ and $p(s|\vee_{i=1}^{n} c_i)$ to be at most $v_2$ with $0 < v_2 < v_1 < 1$. Assume there are $k$ science students in every one of $n$ classes, and there is one arts student in each class, and no arts strudent is in two of the $c_i$. Choose $k \geq v_1/(1-v_1)$ and $n \geq k(1-v_2)/v_2$ and we obtain the desired result.

This means we that for any interpretation of defaults as high probabilities $1-\epsilon$, we can create a counterexample.

### 3.3 "This example is contrived"

It is just a matter of time before someone comes up with a better one; from the 1980 "nonmonotonic logic" special issue of the AI journal to the discovery of the lottery paradox in nonmonotonic logic by Kyburg [1988] was only eight years. The next section describes actual instances of the paradox.

Pearl [1989] argues that these counterexamples are unimportant in most domains. This may be true, but we argue for *testable* and *sound* formalisms that eliminate unwanted inferences even if it means that certain apparently desirable inferences are lost.

## 4 Discussion of the paradox

This variation of the "multiple extension problem" is closely related to Simpson's [1951] paradox of probability theory, (which some think should be attributed to Yule [1903]), which may be stated in a number of ways. Commonly (and perhaps most surprisingly) it is the situation that happens when the truth of $c$ is *known, whether true or false*, then $b$ makes $a$ more probable, but if the truth of $c$ is unknown then $b$ makes $a$ less probable. Formally, it is the fact that there is a consistent assignment of probability values so that for propositions $a$, $b$ and $c$

$$p(a|bc) > p(a|c)$$
$$p(a|b\neg c) > p(a|\neg c)$$
$$p(a|b) < p(a).$$

This situation occurs in real life; Wagner [1982] gives several examples. Possibly the most well-known, though not complete, instance of the paradox was a study of sex bias in graduate admissions at UCB [Bickel et al,1975]. An example can be constructed where admissions by college are fair though campus wide statistics indicate women need higher marks to gain admission. This occurs if most women apply to the most competitive colleges.

Blyth [1973] states that these inequalities are closely related to the facts shown by Chung [1942] that for propositions $a$, $b$ and $c$ there is a consistent assignment of probabilities such that either

$$p(a|b) > p(a)$$
$$p(a|c) > p(a)$$
$$p(a|bc) < p(a)$$

or

$$p(a|b) > p(a)$$
$$p(a|c) > p(a)$$
$$p(a|b \vee c) < p(a).$$

It is a straightforward consequence of the disjunction rule that *both* cannot occur at once:

**Proposition 2** If $a$ is favoured separately by $b$ and $c$, then $a$ is favoured by either $bc$ or $b \vee c$.

**Proof:** From the premises $p(a|b) > p(a)$ and $p(a|c) > p(a)$. Multiplying by the prior of the antecedent and dividing by the prior of the consequent yields $p(b|a) > p(b)$ and $p(c|a) > p(c)$. From the disjunction rule

$$p(bc|a) + p(b \vee c|a) = p(b|a) + p(c|a) > p(b) + p(c) = p(bc) + p(b \vee c).$$

Suppose $a$ does not favour either $bc$ or $b \vee c$. By a similar manipulation, $p(bc|a) \leq p(bc)$ and $p(b \vee c|a) \leq p(b \vee c)$, and

$$p(bc|a) + p(b \vee c|a) \leq p(bc) + p(b \vee c),$$

a contradiction. □

In these examples, $a$ may be conditioned on nine different antecedents: empty antecedent, $b$, $\neg b$, $c$, $\neg c$, $bc$, $b\neg c$, $\neg bc$, $\neg b\neg c$. An ordering on these conditional probabilities must be constrained by the following:

1. for any $a$, $b$ and $c$, $p(a|bc) > p(a|c) > p(a|\neg bc)$ (the direction of the inequality may be reversed),

2. for any $a$, $b$ and $c$, $p(a|c) + p(b|c) = p(ab|c) + p(a \vee b|c)$,

3. if $p(a|b) \neq p(a)$ and $p(a|c) \neq p(a)$, then some combination of outcomes of $b$ and $c$ must increase belief in $a$ and some combination must decrease belief in $a$.

The looseness of the constraints suggests that there are many ways to order the probabilities. Given so many orderings, we should be surprised if there were no surprises.

## 5 Conclusions

Default logic and its variants were proposed as solutions to the problem of reasoning in uncertain domains when numeric probability distributions are unavailable. Few would disagree that such formalisms are becoming awkward even for small problems. We show elsewhere that a system based on probability and the ideas of favouring and of conditional independence seems to yield the expected answers to most of the problems in the nonmonotonic literature. See [Neufeld,1989b,Neufeld and Poole,1988] for details. The most important point relevant to this discussion is that such a system does not in general favour $a$ given $b \vee c$ even though both disjuncts favour $a$.

These results can be interpreted in a number of different ways:

1. The multiple extension problem must be discussed for disjunctive knowledge. The question of how the "typical" individual in $a \vee b$ is chosen must be answered. This means that the meaning of "typical" must be specified and it will be interesting to see if this can be done without introducing a notion of randomness from probability theory.

2. Either it is sensible or not to draw different conclusions from disjunctions than from disjuncts. This obviously varies from one domain to another, but the nature of this variation must be made precise. Perhaps the differences between default logics correspond to statistical properties of different domains.

3. Probability theory tells us that we are taking a chance of being wrong even when the odds are in our favour. The "arts and science" example shows us that some default representations will tell us to "jump" to a conclusion when the odds are *against* us at the outset.

We sum up with a quotation from Koopman [1940] on the same foundational issue: "*The distinction between an asserted disjunction and a disjoined assertion is fundamental: $(u \vee v) = 1$ must never be confused with $(u = 1) \vee (v = 1)$. The disregard of this distinction has led to more difficulties in the foundations of probability than is often imagined.*"

## Acknowledgements

This work resulted in part from years of discussion with David Poole, Romas Aleliunas, and other members of the Logic Programming and Artificial Intelligence Group at Waterloo. Thanks for Dr. W. Knight at the University of New Brunswick for references to Simpson's paradox.

During the refereeing process, an anonymous referee pointed out that Ron Loui [1988] makes similar observations in his thesis. Loui also observes that Simpson's paradox (proof by cases in default logic) is at odds with "specificity".